\documentclass[10pt,twocolumn,letterpaper]{article}

\usepackage{cvpr}              

\usepackage[accsupp]{axessibility}

\usepackage{graphicx}
\usepackage{amsmath}
\usepackage{amssymb}
\usepackage{booktabs}
\usepackage{times}
\usepackage{epsfig}
\usepackage{algorithmic}
\usepackage{algorithm}
\usepackage{caption}
\usepackage{float} 
\usepackage{subcaption}
\usepackage{url}
\usepackage{verbatim}
\usepackage{cite}
\usepackage{textcomp}
\usepackage{multirow}
\usepackage{multicol}
\usepackage{longtable}
\usepackage{tabularx}
\usepackage{makecell}
\usepackage{textcomp}
\usepackage{array} 
\usepackage{longtable}
\usepackage{float}

\newcommand*{\affaddr}[1]{#1} 
\newcommand*{\affmark}[1][*]{\textsuperscript{#1}}

\usepackage[pagebackref,breaklinks,colorlinks]{hyperref}

\usepackage[capitalize]{cleveref}
\crefname{section}{Sec.}{Secs.}
\Crefname{section}{Section}{Sections}
\Crefname{table}{Table}{Tables}
\crefname{table}{Tab.}{Tabs.}

\def\cvprPaperID{7} 
\def\confName{CVPR}
\def\confYear{2023}

\begin{document}

\title{Wild Face Anti-Spoofing Challenge 2023: Benchmark and Results}  

\author{Dong Wang \affmark[1] \qquad Jia Guo  \affmark[2] \qquad Qiqi Shao  \affmark[1] \qquad Haochi He \affmark[1] \qquad  Zhian Chen \affmark[1] \qquad \\
Chuanbao Xiao \affmark[1] \qquad 
Ajian Liu \affmark[3] \qquad 
Sergio Escalera \affmark[4,5] \qquad 
Hugo Jair Escalante \affmark[6] \qquad \\
Zhen Lei \affmark[3] \qquad  
Jun Wan \affmark[3] \qquad 
Jiankang Deng \affmark[2]\thanks{Corresponding author} \\ 
\affaddr{\affmark[1]MoreDian} \qquad
\affaddr{\affmark[2]InsightFace} \qquad
\affaddr{\affmark[3]CASIA}  \qquad\\
\affaddr{\affmark[4]Computer Vision Center (UAB)}  \qquad
\affaddr{\affmark[5]University of Barcelona}  \qquad
\affaddr{\affmark[6]INAOE}  \qquad
\\
{\tt\small \{wangdong,shaoqiqi,hehc,chenzhian,xiaocb\}@moredian.com, }
{\tt\small \{guojia,jiankangdeng\}@gmail.com}\\
{\tt\small sescalera@cvc.uab.cat, }
{\tt\small hugojair@inaoep.mx, }
{\tt\small \{ajian.liu,zhen.lei,jun.wan\}@ia.ac.cn}}

\maketitle

\begin{abstract}
Face anti-spoofing (FAS) is an essential mechanism for safeguarding the integrity of automated face recognition systems. Despite substantial advancements, the generalization of existing approaches to real-world applications remains challenging. This limitation can be attributed to the scarcity and lack of diversity in publicly available FAS datasets, which often leads to overfitting during training or saturation during testing.
In terms of quantity, the number of spoof subjects is a critical determinant. Most datasets comprise fewer than 2,000 subjects. With regard to diversity, the majority of datasets consist of spoof samples collected in controlled environments using repetitive, mechanical processes. This data collection methodology results in homogenized samples and a dearth of scenario diversity.
To address these shortcomings, we introduce the Wild Face Anti-Spoofing (WFAS) dataset, a large-scale, diverse FAS dataset collected in unconstrained settings. Our dataset encompasses 853,729 images of 321,751 spoof subjects and 529,571 images of 148,169 live subjects, representing a substantial increase in quantity. Moreover, our dataset incorporates spoof data obtained from the internet, spanning a wide array of scenarios and various commercial sensors, including 17 presentation attacks (PAs) that encompass both 2D and 3D forms. This novel data collection strategy markedly enhances FAS data diversity.
Leveraging the WFAS dataset and Protocol 1 (Known-Type), we host the Wild Face Anti-Spoofing Challenge at the CVPR2023 workshop. Additionally, we meticulously evaluate representative methods using Protocol 1 and Protocol 2 (Unknown-Type). Through an in-depth examination of the challenge outcomes and benchmark baselines, we provide insightful analyses and propose potential avenues for future research. 
The dataset is released under Insightface\footnote{\href{https://github.com/deepinsight/insightface/tree/master/challenges/cvpr23-fas-wild}{https://github.com/deepinsight/insightface/tree/master/challenges/cvpr23-fas-wild}}.
\end{abstract}

\section{Introduction}
In recent years, face recognition technologies~\cite{arcface,deng2020sub,VPL,zhu2021webface260m,an2022killing} have become increasingly pervasive in various aspects of our lives, including access control, phone unlocking, digital payment, and attendance systems. Despite their widespread adoption, these systems remain susceptible to significant security risks. For instance, face recognition attendance systems are designed to enhance management efficiency; however, if these systems can be easily deceived by photographs, it would result in disarray in personnel management. Similarly, phone unlocking systems, if successfully attacked, may compromise user data security and tarnish the reputation of manufacturers. Malicious attackers may also exploit vulnerabilities in digital payment systems to steal others' identities for illegal purposes. Consequently, face anti-spoofing (FAS) technologies~\cite{wan2020multi,liu2021face,li20203dpc,liu2021cross,liu20213d,liu2019multi,CDCN}, a crucial component of automatic face recognition systems, has garnered considerable attention from both academia and industry.

Owing to the rapid progress of deep learning, the FAS technology community has witnessed a surge in outstanding contributions. Deep learning-based FAS methods can be categorized into three groups: 1) classification supervision, 2) auxiliary pixel-wise supervision, and 3) generative pixel-wise supervision. Intuitively, FAS tasks can be framed as classification problems, with many works~\cite{TSCNN,SSR-FCN,DRL-FAS,PCGN,PatchNet,ViTranZFAS} directly supervised using binary cross-entropy (CE), while others~\cite{FGDNet,100} extend the problem to multiple classification. Classification supervisions are easy to construct, enabling deep FAS models to converge rapidly. In contrast, pixel-wise supervision with auxiliary tasks can extract more fine-grained cues, with additional information such as pseudo depth maps~\cite{CDCN,DCCDN,TransFAS,DepthSeg}, binary mask maps~\cite{DTN,BCN,MPFLN}, and reflection maps~\cite{DCN, CelebA-Spoof,BASN} helping to delineate local live/spoof features. Generative pixel-wise supervisions~\cite{baidu-spoofcue,STDN,GOGen,PhySTD,MT-FAS}, which do not rely on expert-designed guidance and offer more flexible labels, visualize spoof cues in spoof samples, thereby enhancing the interpretability of FAS tasks.

As highlighted in~\cite{CelebA-Spoof}, existing FAS methods continue to face challenges in generalizing to real-world scenarios, particularly when employing unimodal RGB sensors without hardware advantages. These sensors are prone to various presentation attacks (PAs), including print, replay, and 3D-model attacks. In comparison to face recognition technology, face anti-spoofing remains an unresolved issue in face recognition systems~\cite{Zitong-Yu-A-Survey,wan2021special}. The success of recognition technology largely depends on the availability of large-scale, diverse datasets. The 2016 release of the MS1M~\cite{MS1M} dataset marked a turning point in the rapid development and industrial application of face recognition algorithms.

The relatively slow progress of face anti-spoofing technology can be attributed to the limitations in the quantity and diversity of publicly accessible FAS datasets, which often leads to overfitting during training or saturation during testing. FAS datasets typically comprise several key components: the number of subjects, presentation attacks (PAs), scenarios, and input sensors. The scale of a dataset is primarily determined by the number of subjects; however, most existing FAS datasets include fewer than 2,000 subjects, with only one dataset (CelebA-Spoof) containing over 10,000 subjects. Moreover, CelebA-Spoof has an average of more than 60 images per subject, which results in high data homogeneity and adversely impacts dataset diversity.

In terms of diversity, the remaining elements play a crucial role. PAs can be broadly classified into 2D and 3D forms based on their geometric properties. 2D PAs display facial identity information to sensors through photos or videos, with common attack variants including flat or wrapped printed photos, cut-out photos, images displayed on screens, and video replays. With advancements in 3D manufacturing technology, 3D masks and models have emerged as new PAs that challenge FAS technology. In comparison to traditional 2D PAs, 3D attacks exhibit greater realism in terms of texture and geometric structure. Rigid 3D masks can be made from various materials, such as paper, resin, plaster, or plastic, while flexible soft 3D masks typically consist of silicone or latex. 3D models often exhibit high levels of simulation, including waxworks and adult dolls.

A review of current datasets, as shown in Table~\ref{datasets}, reveals that most datasets contain either 2D or 3D PAs exclusively. Datasets such as \cite{NUAA, Msspoof,zhang2020casia, UVAD, REPLAYMobile, REPLAYATTACK, MSUMFSD, MSUUSSA, OULUNPU} feature 2D print or display PAs, while \cite{HKBUMARs-V2,yu2020fas, liu2022contrastive, 3DMAD, [a8], [a10], WFFD, [SWAX]} include 3D mask or model PAs. Some datasets, such as \cite{Rose-Youtu, DTN, CelebA-Spoof}, encompass both 2D and 3D PAs; however, their diversity remains unsatisfactory due to deficiencies in other elements, such as the absence of high-fidelity 3D models.
Additionally, the spoof samples in nearly all current datasets are collected in controlled and limited scenarios through mechanical and repetitive processes, which we refer to as manually controlled scenarios. This data collection approach results in a lack of scenario richness and leads to significant sample homogenization.

To address the limitations of existing face anti-spoofing datasets, we introduce the Wild Face Anti-Spoofing Dataset (WFAS), a large-scale FAS dataset collected in the wild. To the best of our knowledge, this is the first dataset to extend FAS research to real-world scenarios. Our dataset comprises 853,729 images of 321,751 spoof subjects and 529,571 images of 148,169 live subjects, significantly increasing the quantity of available data. Furthermore, the spoof data is sourced from the internet, encompassing a wide variety of scenarios and commercial sensors.

The internet-derived spoof samples, though not intended to attack face recognition systems, incidentally resemble spoof samples that benefit FAS research. Examples include 2D faces appearing in picture books or on TV screens, as well as 3D waxwork faces at tourist attractions. Our dataset includes 17 PAs, covering both 2D and 3D forms. The 2D PAs comprise print types (\eg, newspapers, posters, photos, albums, picture books, scanned photos, packaging, and cloth) and display types (\eg, phones, tablets, TVs, and computers). The 3D PAs feature five subcategories with varying fidelity: masks, garage kits, dolls, adult dolls, and waxworks.

\begin{figure*}[htbp]    
    \centering
    \includegraphics[width=0.9\linewidth]{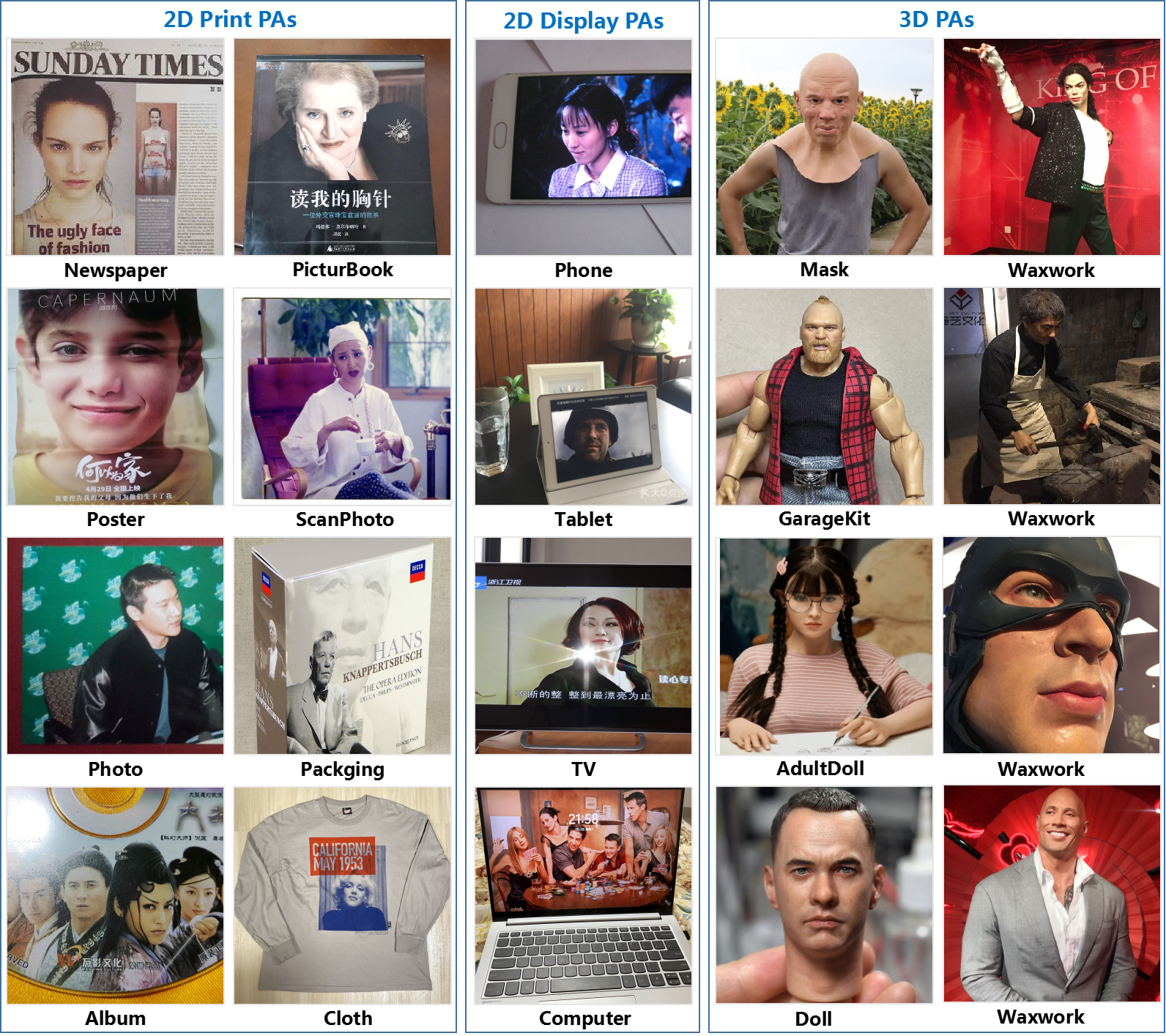}
    \caption{Examples of spoof faces in our dataset.}
    \label{fig:patch}
\end{figure*}

All spoof samples were captured using photographic equipment, such as various mobile phone brands, digital cameras, and scanners. These optical sensors produce images with a wide range of resolutions, resulting in a richer face quality within our dataset. The live face set, collected from the internet under specific creative commons licenses, is a typical in-the-wild dataset, incorporating diverse scenarios, races, ages, and more. Our dataset significantly enhances FAS data diversity, as illustrated by the spoof samples in Figure~\ref{fig:patch}.

Leveraging our dataset and Protocol 1 (Known Type), we hosted the Wild Face Anti-Spoofing Challenge at the CVPR2023 workshop. The competition attracted 219 teams, with 66 teams advancing to the final round. The top-ranking algorithms were re-run and analyzed by the organizing team. We also thoroughly benchmarked existing representative methods on Protocol 1 and Protocol 2 (Unknown Type). Based on a comprehensive examination of the challenge results and benchmark baselines, we provide insightful analysis and discuss future research directions.

\renewcommand\arraystretch{1}
\begin{table*}
\centering
\caption{Face anti-spoofing datasets recorded by commercial RGB cameras.}
	\label{datasets}
\scalebox{0.85}{
 \begin{tabular}{p{3.5cm}<{\centering}p{1.5cm}<{\centering}p{1.5cm}<{\centering}p{2.5cm}<{\centering}p{2.5cm}<{\centering}p{6.5cm}<{\centering}}
  \toprule
   
        Dataset                       &    Year    &    Subjects     &    Quantity   & Format  & PAs \\ 
        \midrule
        NUAA\cite{NUAA}               &    2010    &      15         &    12,614   & image   & Print\\
        PRINT-ATTACK\cite{[a3]}       &    2011    &      50         &     400     & video     & Print\\
        CASIA-FASD\cite{[a2]}         &    2012    &      50         &     400     & video    & Print, Replay\\
        REPLAY-ATTACK\cite{REPLAYATTACK}    &    2012    &      50   &    1,200    & video    & Print, Replay\\
        3DMAD\cite{3DMAD}              &    2014    &      17         &     255    & video    & Mask(paper, hard resin) \\
        MSU-MFSD\cite{MSUMFSD}         &    2014    &      35         &     440    & video      & Print, Replay\\
        Msspoof\cite{Msspoof}          &    2015    &      21         &    4,704 & image      & Print \\
        UVAD\cite{UVAD}             &    2015    &     404         &    17,076    & video     & Replay \\
        MSU-USSA\cite{MSUUSSA}         &    2016    &     1,140        &    10,260 & image    & Print, Replay \\
        REPLAY-Mobile\cite{REPLAYMobile}    &    2016    &      40         &    1,030    & video      & Print, Replay\\
        3DFS-DB\cite{[a8]}            &    2016    &      26         &    520    & video       & Mask(plastic)\\
        HKBU-MARs V2\cite{[a9]}       &    2016    &      12         &    1,008    & video      & Mask(hard resin) \\
        BRSU\cite{[a10]}              &    2016    &      6          &    140    & video      & Mask(silicon, plastic, resin, latex) \\ 
        OULU-NPU\cite{OULUNPU}          &    2017    &      55         &    3,600    & video      & Print, Replay \\ 
        Rose-Youtu\cite{Rose-Youtu}  &    2018    &      20         &    3,350    & video     & Print, Replay, Mask(paper, crop-paper) \\ 
        WFFD\cite{[WFFD]}             &    2019    &     745         &  4,600/285    & image/video & Waxworks(wax) \\ 
        SiW-M\cite{DTN}             &    2019    &     493         &    1,628    & video      & Print, Replay, Mask(hard resin, plastic, silicone, paper, Mannequin) \\
        CASIA-SURF\cite{zhang2020casia}        &    2019    &     1,000        &    21,000     & video    & Print\\ 
        SWAX\cite{[SWAX]}             &    2020    &      55         &  1,812/110    & image/video & Waxworks(wax) \\
        CelebA-Spoof\cite{CelebA-Spoof}      &    2020    &     10,177       &    625,537 & image    & Print, Replay, Mask(paper) \\
        CASIA-SURF 3DMask\cite{yu2020fas}  &    2020    &      48         &    1,152    & video      & Mask(3D print) \\
        CASIA-SURF CeFA\cite{liu2021casia}   &    2021    &     1,607        &    23,538      & video  & Print, Replay, Mask(3D print, silica gel) \\
        HiFiMask\cite{liu2022contrastive}          &    2021    &      75         &    54,600    & video     & Mask(transparent, plaster, resin) \\
        \textbf{\textbf{Our Dataset (WFAS)}}       &    2023    &      469,920         &    1,383,300& image     & Print(newspaper, poster, photo, album, picture book, scan photo, packging, cloth), Display(phone, tablet, TV, computer), Mask, 3D Model(garage kit, doll, adult doll, waxwork)        \\
  \bottomrule
  \end{tabular}}
\end{table*}

\section{Related Work}
\subsection{Face Anti-Spoofing Datasets}
As shown in Table~\ref{datasets}, we focus on datasets recorded using commercial RGB cameras. The first dataset specifically designed for the face anti-spoofing field is the NUAA Photograph Imposter Database (NUAA)\cite{NUAA}, containing only 2D print attack types with 15 subjects and 500 images per subject. In\cite{[a2]}, the authors introduced CASIA-FASD, featuring three types of PAs: distorted printed photos, printed photos with perforated eye areas, and video replays. This dataset can be seen as an extension of NUAA, increasing PA diversity.

PRINT-ATTACK~\cite{[a3]} was the first dataset to provide accurate protocols, including training, evaluation, and testing sets, and contains attack videos of 50 printed photos with different identities. REPLAY-ATTACK~\cite{REPLAYATTACK} added more PA types, established a protocol for fair comparison of face anti-spoofing algorithms, and demonstrated the vulnerability of face recognition systems to these attacks. Additional similar 2D print or replay datasets include MSU-MFSD~\cite{MSUMFSD}, MSU-USSA~\cite{MSUUSSA}, Msspoof~\cite{Msspoof}, UVAD~\cite{UVAD}, and REPLAY-Mobile~\cite{REPLAYMobile}.

Compared to 2D PAs, 3D masks offer a more realistic texture and geometric structure, making them more effective at deceiving FAS systems. 3DMAD~\cite{3DMAD} was the first published 3D mask FAS dataset, featuring 255 videos of 17 subjects. The rigid mask is made of paper and hard resin. Subsequent datasets like 3DFS-DB~\cite{[a8]}, HKBU-MARs V2~\cite{[a9]}, and BRSU~\cite{[a10]} improved acquisition equipment, mask types, and lighting environments. WFFD~\cite{[WFFD]} introduced the first waxwork dataset, comprising 450 subjects and 2,200 images, significantly increasing fidelity. Similarly, SWAX~\cite{[SWAX]} included images and videos of waxworks. Recent 3D FAS datasets have considered new attack types, ethnic diversity, and complex recording conditions, such as SiW-M~\cite{DTN} with 13 fine-grained attack types, CASIA-SURF CeFA~\cite{liu2021casia} addressing ethnic bias with three ethnicities, and HiFiMask~\cite{liu2022contrastive} encompassing six lighting conditions, seven recording devices, and six scenarios.

However, the datasets mentioned above are limited in quantity, particularly in terms of subjects. In this context, the authors of CelebA-Spoof \cite{CelebA-Spoof} explicitly address the issue of FAS dataset scale, expanding the dataset and increasing the number of subjects to over ten thousand to advance the FAS community. CelebA-Spoof \cite{CelebA-Spoof} is currently the largest dataset with 625,537 images of 10,177 subjects across eight scenarios, boasting rich annotations. However, the top three teams achieved TPR=100$\%$@FPR = $5*10^{-3}$ on this dataset in the ECCV2020 FAS challenge~\cite{194}, indicating that dataset diversity cannot be neglected while increasing quantity.

To further promote advancements in the FAS dataset with respect to both quantity and diversity, we introduce the first large-scale FAS dataset in the wild, named the Wild Anti-Spoofing Dataset (WFAS). WFAS comprises 853,729 images featuring 321,751 spoof subjects and 529,571 images of 148,169 live subjects, resulting in a substantial increase in quantity. Furthermore, the spoof data is sourced from the internet, encompassing a broad range of scenarios and various commercial sensors. With 17 PAs covering 2D and 3D forms, this innovative data pattern in the wild represents a significant breakthrough in FAS data diversity.

\subsection{Face Anti-Spoofing Methods}
As CNN architectures develop and FAS datasets are progressively released, end-to-end deep learning-based methods have come to dominate the field of FAS. Deep learning-based FAS methods can be classified into three categories: 1) classification supervision, 2) auxiliary pixel-wise supervision, and 3) generative pixel-wise supervision.

Classification-based methods typically employ binary cross-entropy supervision. In \cite{[a6]}, an end-to-end deep learning FAS method based on CNN structure is proposed for the first time. To mitigate overfitting, \cite{TSCNN, ViTranZFAS} pre-trained the models on ImageNet. To adapt to the low computing performance of mobile and edge platforms, \cite{55} proposed using the lightweight MobileNetV2~\cite{SSR-FCN}. Some research focused on structural optimization; for instance, \cite{132} suggested employing a shallow fully convolutional network (FCN) to construct a multi-scale structure for learning local discriminative cues in FAS. Other work concentrated on loss function optimization. FAS tasks often exhibit asymmetric intra-distributions, with the living class being more compact and the spoof class being more diverse. \cite{PatchNet} introduced an asymmetric angular-margin softmax loss to ease intra-class constraints among PAs. To enhance the prediction of hard samples, binary focal loss was utilized to expand the margin between live/spoof samples, resulting in stronger discrimination for hard samples\cite{138}. However, binary classification models are prone to overfitting and lack robustness to attacks in scenarios with minor domain shifts. Some researchers reframed FAS as a fine-grained classification problem~\cite{FGDNet,100}, in which type labels are defined as bonafide, print, replay, \etc. Despite this, models with multi-class CE loss still struggle to distinguish sample distributions between live and spoof, particularly for hard samples.

Pixel-wise supervision with auxiliary tasks skillfully leverages human prior knowledge. Pseudo depth labels~\cite{CDCN,[c26],[c96],140} take into account that 2D PAs lack facial depth information. These works require deep models to predict genuine depth for live samples while producing zero maps for spoof ones. Pseudo reflection maps~\cite{BCN, BASN, CelebA-Spoof} consider the discernible cues between the reflection of bonafide samples and PAs. The accuracy of pseudo depth/reflection labels depends on the precision of models with other relevant tasks. In contrast to these labels, binary mask labels~\cite{[c32], DTN, STDN, 143, [c145]} are more suitable for PAs with facial depth (\eg, waxworks) and easier to generate. However, binary mask labels used in current methods typically assume that all pixels in the face region have the same live/spoof distributions, generating all ``one'' and ``zero'' maps for bonafide and PAs, respectively. Such labels are imprecise and challenging to learn when dealing with partial attacks (\eg, paper masks with perforated eye areas).

Unlike auxiliary pixel-wise supervision, pixel-wise supervision with generative models~\cite{baidu-spoofcue, STDN, GOGen, PhySTD, MT-FAS, 33, 139} does not impose expert-designed hard constraints. The pixel-wise labels of these generative models are softer, allowing for a wider space for implicit spoof cue discovery. These works typically reframe FAS as a spoof noise modeling problem and design an encoder-decoder architecture to estimate the underlying spoof patterns with relaxed pixel-wise supervision (\eg, zero-noise maps for live faces). Generative pixel-wise supervision is visually insightful and more interpretable. However, such soft pixel-wise supervision may easily become trapped in local optima and overfit to unexpected interference (\eg, sensor noise), resulting in poor generalization under real-world scenarios.

\section{Wild Face Anti-Spoofing Dataset}
\subsection{Data Construction}

\noindent{\bf Live Data.} The live samples in our dataset were obtained from the internet under a specific creative commons license, making it a typical face dataset in the wild that includes a variety of scenarios, races, ages, \etc. All live faces were clustered using RetinaFace~\cite{retinaface} and ArcFace~\cite{arcface}, resulting in a total of 529,571 images from 148,169 live subjects.

\noindent{\bf Spoof Data.} Our spoof data boasts unique characteristics that make the dataset progressive compared to current alternatives. Instead of manual collection, our spoof images were sourced from the internet, encompassing a wide range of scenarios and various commercial sensors. These images were not created to intentionally attack face recognition systems but happened to share similar presentation characteristics to spoof samples, benefiting our FAS research. Examples include 2D faces appearing in picture books or on TV screens and 3D waxwork faces at tourist attractions.

The PAs in our dataset encompass both 2D and 3D forms with diverse materials. The 2D PAs include 2D print and 2D display PAs. The former has four styles (\ie, bending, folding, cutting, and plane) and eight carriers (\ie, newspaper, poster, photo, album, picture book, scanned photo, packaging, and cloth). The latter comprises four subcategories (\ie, phone, tablet, TV, and computer) with screen types including LCD, IPS, OLED, and VA, among others.

3D PAs feature five subcategories with varying levels of fidelity, such as masks, garage kits, dolls, adult dolls, and waxworks, made from materials like resin, plaster, plastic, silicone, latex, and wax. Our dataset contains a total of 17 PAs covering both 2D and 3D forms. Detailed subject and image numbers for all PAs are provided in Table~\ref{details}.

All spoof samples were captured using a range of photographic equipment, including various mobile phone brands, digital cameras, and even scanners. The optical sensors of this equipment captured images with diverse resolutions, resulting in a richer face quality within our dataset. This new data pattern in the wild represents a significant breakthrough in FAS data quantity and diversity. Adopting such an FAS data production method saves considerable labor, material resources, and time.

\renewcommand\arraystretch{1}
\begin{table}
\caption{Dataset details including subjects/images distribution of training/development/testing settings.}
\label{details}
\resizebox{\columnwidth}{!}{
 \begin{tabular}{c|cccccc}
  \toprule
   
        Category                     &    PAs    &    Subjects     &    Images  &Train  &Dev  &Test      \\ 
    \midrule
    \multirow{8}{*}{2D-Print}
                & Newspaper      &  9,046       &  14,425   &  \checkmark  &    &  \\
                & Poster         &  40,858      &  15,439   &  \checkmark  &    &  \\
                & photo          &  61,990      & 102,826   &  \checkmark  &    &  \\
                & Album          &  21,122      & 56,490    &  \checkmark  &    &  \\
                & PictureBook    &  118,355     & 349,232   &  & \checkmark  & \checkmark\\
                & ScanPhoto      &  1,161       & 2,484     &  & \checkmark  & \checkmark\\
                & Packaging      &  3,866       & 19,136    &  & \checkmark  & \checkmark\\
                & Cloth          &  138        & 266      &  & \checkmark  & \checkmark\\
    \midrule
    \multirow{4}{*}{2D-Display}
                & Phone          & 20,813       & 34,907   &  \checkmark  &    &   \\
                & Tablet            & 8,089        & 15,431   &  \checkmark  &    &   \\
                & TV             & 28,184       & 75,606   &  & \checkmark  & \checkmark\\
                & Computer       & 13,938       & 25,291   &  & \checkmark  & \checkmark\\
    \midrule 
    \multirow{5}{*}{3D}
                & Mask           & 268         & 1,454    &  \checkmark  &    &  \\
                & GarageKit      & 1,488        & 4,505    &  \checkmark  &    &  \\
                & AdultDoll      & 165         & 12,021   &  \checkmark  &    &  \\
                & Doll           & 15,406       & 91,954   &  & \checkmark  & \checkmark\\
                & Wax            & 2,283        & 6,843    &  & \checkmark  & \checkmark\\
  \bottomrule
  \end{tabular}}
\end{table}

\renewcommand\arraystretch{1}
\begin{table*}[!ht]
 \centering
 \caption{Baseline performance under Protocol 1 and Protocol 2.}
 \label{tab:SOTA}
 \scalebox{0.9}{
\begin{tabular}{c|c|c|c|c|c|c|c} \hline 
\multicolumn{1}{c|}{\multirow{2}{*}{Prot.}} & \multicolumn{1}{c|}{\multirow{2}{*}{SOTA Method}} & \multicolumn{1}{c|}{\multirow{2}{*}{Flops}}
& \multicolumn{2}{c|}{Dev} & \multicolumn{3}{c}{Test} \\  \cline{4-8} 
 & &  & EER threshold & EER(\%) & APCER(\%) & BPCER(\%) & ACER(\%) \\ \hline
 
  \multirow{8}{*}{1}
 & Res-50 & 4.13G & 0.6736 & 4.55 & 6.45 & 8.96 & 7.71   \\ \cline{2-8}
& PatchNet & 1.82G & 0.8415 & 5.78 & 8.13 & 8.94 & 8.53\\ \cline{2-8}
& MaxVit & 4.46G &0.4964 & 3.57 & 5.44 & 7.72 & \textbf{6.58}\\ \cline{2-8}
& CDCN++binarymask & 79.79G & 0.4374 & 10.96 & 11.15 & 11.84 & 11.50\\ \cline{2-8}
& CDCN++ & 50.97G & 0.2692 & 7.75 & 7.96 & 9.52 & 8.74\\ \cline{2-8}
& DC-CDN & 354.94M & 0.2656 & 8.99 & 9.26 & 11.96 & 10.61\\ \cline{2-8}
& DCN & 49.31G & 0.9915  & 18.58 & 19.93 & 18.88 & 19.40\\ \cline{2-8}
& LGSC & 9.56G &4.56e-5 & 5.32 & 7.60 & 9.39  & 8.50\\ \hline

 \multirow{8}{*}{2}
 & Res-50 & 4.13G & 0.5647 & 4.02 & 47.10 & 7.76 &  27.43  \\ \cline{2-8}
& PatchNet & 1.82G & 0.6957 & 3.99 & 58.74 & 6.07 & 32.40\\ \cline{2-8}
& MaxVit & 4.46G & 0.4516 & 3.13 & 51.50 & 6.74 & 29.12 \\ \cline{2-8}
& CDCN++binarymask & 79.79G & 0.3774 & 10.50 & 49.76 & 11.68 & 30.72\\ \cline{2-8}
& CDCN++ & 50.97G & 0.2406 & 6.85 & 51.69 & 8.52 & 30.11 \\ \cline{2-8}
& DC-CDN & 354.94M & 0.2559 & 9.90 & 54.86 & 11.69 & 33.28 \\ \cline{2-8}
& DCN & 49.31G & 0.9877 & 14.40 & 15.71 & 68.63 & 42.17 \\ \cline{2-8}
& LGSC & 9.56G & 1.91e-4 & 4.65 & 45.38 & 7.55 & \textbf{26.47} \\ \hline
\end{tabular}}
\end{table*}

\section{Experimental Settings}
The dataset is divided into training, development, and testing subsets with an approximate ratio of 4:1:5. As shown in Table~\ref{details}, each PA does not appear in the same subset simultaneously. For instance, some PAs (\eg, Newspaper, Poster, Photo, and Album) belonging to the 2D Print category appear only in the training set, while others are found in the development or testing set. This arrangement leverages the advantages of data diversity and examines the algorithm's robustness in cases of slight domain shifts.

\subsection{Evaluation Metrics}
We adopt the widely used standardized ISO/IEC 30107-3 metrics for performance evaluation. The evaluation metric comprises Attack Presentation Classification Error Rate (APCER), Bonafide Presentation Classification Error Rate (BPCER), and Average Classification Error Rate (ACER). APCER and BPCER measure the error rates of spoof and live samples, respectively. ACER is calculated as the mean of APCER and BPCER. The formulas are as follows:
\begin{equation}
APCER = FP/(TN+FP),
\end{equation}
\begin{equation}
BPCER = FN/(TP+FN),
\end{equation}
\begin{equation}
ACER = (APCER+BPCER)/2,
\end{equation}
where FN, FP, TN, and TP represent false negatives, false positives, true negatives, and true positives, respectively. Additionally, Equal Error Rate (EER)~\cite{[a67]} is used in the development set to obtain the threshold, which is then employed to calculate APCER and BPCER in the testing set.

\subsection{Evaluation Protocols}
\noindent{\bf Known-Type Protocol.} In contrast to the widely used intra-dataset and intra-type protocol, which evaluates each PA type individually, we introduce a new protocol called Known-Type Protocol. This protocol employs all PA types for training, development, and testing, offering a more global, compatible, and realistic application scenario.

\noindent{\bf Unknown-Type Protocol.} This protocol designates one PA type to appear exclusively in the testing stage to assess whether the algorithms have learned generalized spoof cues for unknown attack types. In this work, 2D PAs are utilized in the training and development stages, while 3D PAs are employed in the testing stage.

\subsection{Baselines}
We evaluate various representative algorithms on Protocol 1 and Protocol 2, including classification supervision (\ie, ResNet-50~\cite{resnet}, PatchNet~\cite{PatchNet}, MaxVit~\cite{maxvit}), auxiliary pixel-wise supervision (\ie, CDCN++\cite{CDCN}, CDCN++binarymask\cite{CDCN++_binary}, DCN\cite{DCN}, DC-CDN\cite{DCCDN}) and generative pixel-wise supervision (\ie, LGSC\cite{baidu-spoofcue}).
For a fair comparison, we do not use pre-trained models, and other parameters are reproduced according to the original works.
Additionally, to eliminate the influence of arbitrary and unfaithful cues (\eg, screen bezel) on spoofing patterns, we use the face bounding box~\cite{retinaface} as the input scale, forcing the models to focus on the face area for feature learning and prediction.
The results are listed in Table~\ref{tab:SOTA}, where we find that classification supervision-based models perform better in both Protocol 1 and Protocol 2. The performances of models with greater designability (\ie, auxiliary pixel-wise supervision) fall short of expectations. Generative pixel-wise supervision significantly outperforms auxiliary pixel-wise supervision in Protocol 1 and demonstrates exceptional potential in Protocol 2, achieving the top performance.

\renewcommand\arraystretch{1}
\begin{table*}[t]
 \centering
 \caption{The top-10 results of the Wild Face Anti-Spoofing Challenge at CVPR2023 workshop.}
 \label{competition}
 \scalebox{1.0}{
\begin{tabular}{c|c|c|c|c|c} \hline 

Rank  & Affiliation & Team & ACER(\%) & APCER(\%) & BPCER(\%)  \\ \hline
 
1  & ChinaTelecom & xuyaowen & {\bf 1.6010} & {\bf 1.2960} & {\bf 1.9060}  \\ \hline
2  & Meituan & hexianhua & 2.2210 & 1.3770 & 3.0640  \\ \hline
3  & Netease & buccellati & 2.5540 & 2.3390 & 2.7690  \\ \hline
4  & SCUT & xmj & 2.8940 & 1.4440 & 4.3450  \\ \hline
5  & - & luoman & 3.0700 & 1.7450 & 4.3950  \\ \hline
6  & - & Sicks & 3.1450 & 1.7250 & 4.5640  \\ \hline
7  & KiwiTech & KiwiTech\_LeoDu & 3.1800 & 2.2060 & 4.1540   \\ \hline
8  & XMU & Iverson & 3.1890 & 3.2890 & 3.0900  \\ \hline
9  & - & admin123 & 3.5300 & 2.7530 & 4.3060  \\ \hline
10 & SJTU & iKun\-CTRL & 3.5430 & 3.2420 & 3.8440  \\ \hline

\end{tabular}}
\end{table*}

\subsection{Challenge Results}
\begin{figure}[htbp]    
    \centering
    \includegraphics[width=1.0\linewidth]{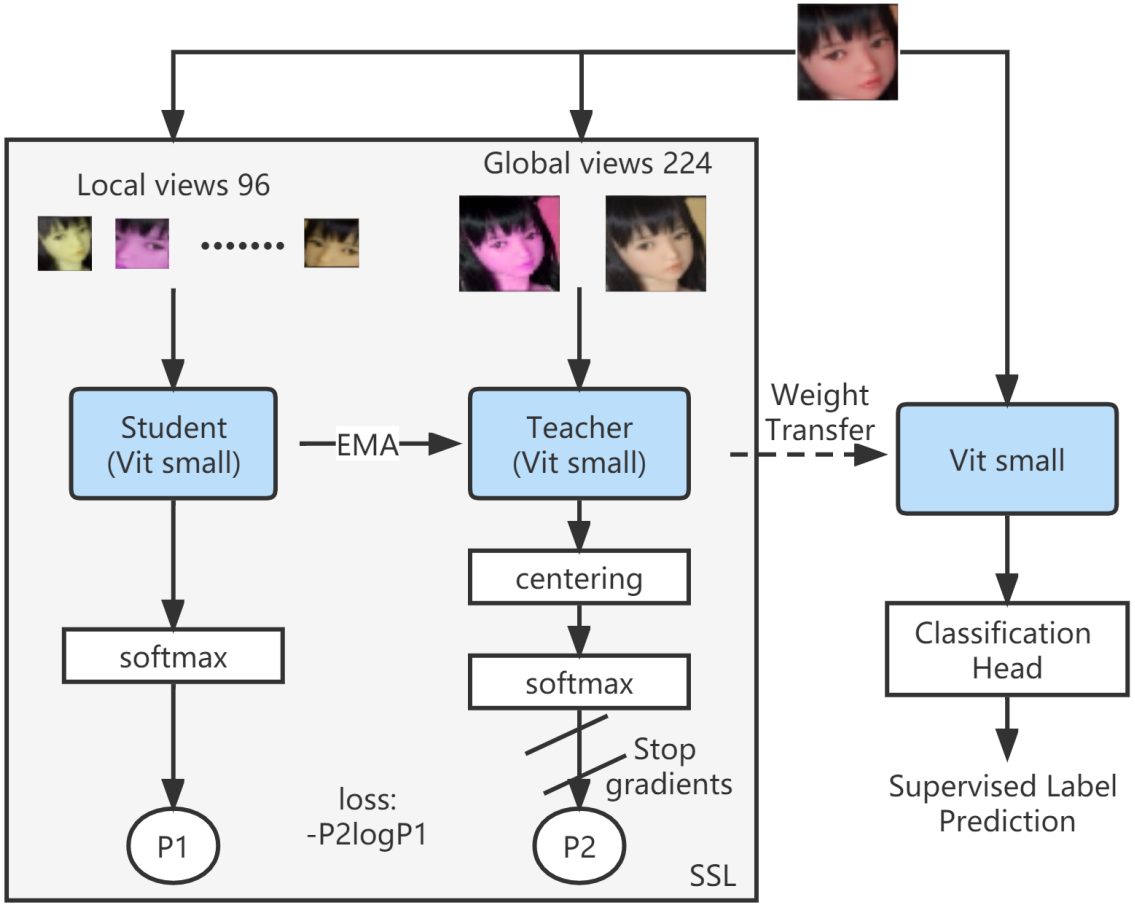}
    \caption{Method diagram of ChinaTelecom.}
    \label{ChinaTelecom}
\end{figure}

Based on our dataset and Protocol 1 (Known-Type), we hosted the Wild Face Anti-Spoofing Challenge at the CVPR 2023 workshop. A total of 219 teams participated in the competition, with 66 teams advancing to the final round. The top-ranking algorithms were re-run and analyzed by the organizing team. Table~\ref{competition} presents the top-10 results of the Wild Face Anti-Spoofing Challenge at the CVPR 2023 workshop. In contrast to the baseline evaluation, we relaxed the face scale constraint for the Challenge. The face scale was randomly set to between 1.0 and 1.2 times the side length, making it more suitable for real-world application scenarios (\eg, exposing part of the screen bezel or book edge). Ultimately, the teams from China Telecom, Meituan, and NetEase secured the top three positions in the competition. Here is a brief introduction to their solutions:

\noindent{\bf China Telecom.} As illustrated in Figure~\ref{ChinaTelecom}, the champion solution consists of two learning phases, \ie, self-supervised \cite{telecom[3]_self-supervised} stage and supervised learning stage. In the first stage, the model passes two different views with random transformations of an input image, without labels, to the student and teacher networks. Both networks have the same architecture but different parameters. Each network outputs a K-dimensional feature that is normalized with a temperature softmax over the feature dimension. The output of the teacher network is centered with a mean computed over the batch. Their similarity is then measured with a cross-entropy loss. The stop-gradient \cite{telecom[2]_stop-gradient} operator is applied to the teacher to propagate gradients only through the student network. The teacher network's parameters are updated with an exponential moving average (EMA) of the student parameters.

In the second stage, self-supervised weights are used to initialize ViT \cite{telecom[4]_ViT} for supervised learning. Meanwhile, a series of data augmentation strategies (\eg, color jitter, clahe, Gaussian blur, random fog, random crop, random flip) are adopted to improve the generalization ability. The AdamW \cite{telecom[1]_AdamW} optimizer is used for 20 epochs of training, and the initial learning rate is set to 2e-6. The CosineAnnealingLR \cite{telecom[5]_CosineAnnealingLR} is adopted to reduce the learning rate, and cross-entropy loss is used for supervised learning.

\begin{figure}[htbp]    
    \centering
    \includegraphics[width=1.0\linewidth]{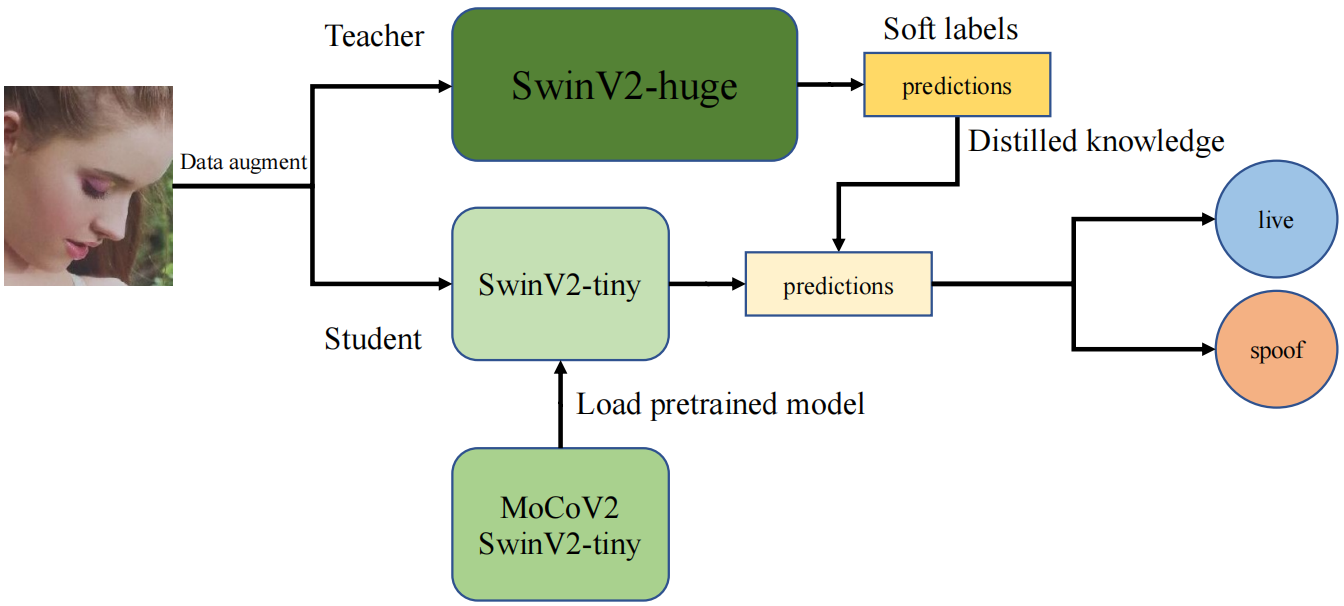}
    \caption{Method Diagram of MeiTuan.}
    \label{MeiTuan}
\end{figure}

\noindent{\bf MeiTuan.} Firstly, the unlabeled training and development sets are merged to form the self-supervised data. The MoCoV2 \cite{meituan[1]} self-supervised method is then employed to train two pre-trained models, \ie, SwinV2-huge and SwinV2-tiny \cite{meituan[2]}. These pre-trained models are fine-tuned for the face anti-spoofing task, resulting in significant performance improvements. Secondly, the team utilizes the knowledge distillation method DKD \cite{meituan[3]} to further enhance the performance of the face anti-spoofing model. They use SwinV2-huge as the teacher model and SwinV2-tiny as the student model. Thirdly, appropriate data augmentation techniques are applied to improve the model's generalization. The main process of the method is shown in Figure~\ref{MeiTuan}.

\noindent{\bf NetEase.} The team from NetEase proposes a two-stage training strategy to improve the model's capability without pre-training. First, they train a convnext \cite{convnext_V2} model on the training set based on binary cross-entropy loss and then use the model to infer the soft labels of the training set. Second, they construct a compound loss using focal loss and triplet loss, and then fit the soft labels with the maxvit \cite{maxvit} model. Different input formats are used for the two stages of model training. For convnext, the field of view is fixed at 1.2 times the face box's outer expansion, while maxvit employs random outer expansion during training. They observe that maxvit learns more about face anti-spoofing from the soft labels of convnext. Additionally, data enhancements such as mixup, cutmix, and color adjustment are utilized to improve the model's generalization.

\section{Conclusion and Discussion}
In this paper, we construct a large-scale FAS dataset in the wild, named Wild Face Anti-Spoofing (WFAS) Dataset. The WFAS Dataset represents a significant breakthrough in FAS data quantity and diversity, and most importantly, it paves the way for a vast scale of wild FAS data in the future. Based on our dataset and Protocol 1, we host the Wild Face Anti-Spoofing Challenge at the CVPR2023 workshop and analyze the top-ranking algorithms. Moreover, we thoroughly benchmark representative methods on Protocol 1 and Protocol 2.

An analysis of the baseline results reveals that state-of-the-art algorithms in recent years do not demonstrate robustness to changes in scenarios (\ie, dataset changes). Interestingly, the simpler classification supervision-based method achieved better results. This raises the question of whether current models with complex designs can effectively mine intrinsic FAS features. We believe that generative pixel-wise supervision methods, which offer more interpretability for visual spoof patterns, have greater potential for future developments, as evidenced by their performance across both protocols in our large-scale WFAS dataset.

Additionally, upon further investigation, we find that most current generative pixel-wise supervision methods focus on the spoof class, such as spoof noise modeling or spoof cue generation. FAS tasks typically exhibit asymmetric intra-distributions, with the live class being more compact and the spoof class being more diverse. As a result, we argue that the current definition of generative pixel-wise supervision revolving around the spoof class lacks rationality, since the PAs of the spoof class are constantly evolving and expanding. Minimizing compact live cues in spoof samples is a more reasonable approach than minimizing diverse spoof cues in live samples.

In reviewing the challenge, the top-3 solutions all utilized the Transformer \cite{2017Attention} architecture, which has been proven to outperform CNN in several tasks. Notably, two of the winners applied self-supervised learning to build pre-trained models as a solid initialization for FAS tasks, which was one of the key factors for their success. This approach addresses the problem of the Transformer architecture's difficulty converging in computer vision tasks. Although the competition results surpassed the baseline results, we have yet to see the emergence of new methods with better interpretability for the FAS task.

{\small
\bibliographystyle{ieee_fullname}
\bibliography{egbib}
}

\end{document}